\pdfoutput=1

\documentclass[11pt]{article}
\usepackage{booktabs}
\usepackage[]{acl}
\usepackage{comment}

\usepackage{amsmath}

\usepackage{times}
\usepackage{latexsym}
\usepackage{comment}
\usepackage[T1]{fontenc}

\usepackage[utf8]{inputenc}

\usepackage{microtype}
\usepackage{graphicx}
\usepackage{xcolor}

\newcommand{\trisha}[1]{\textcolor{red}{#1}}

\title{Synthetic Patient-Physician Dialogue Generation from Clinical Notes Using LLM}

\author{
  Trisha Das\thanks{Equal contribution},
  Dina Albassam\footnotemark[1],
  Jimeng Sun\\
  \texttt{trishad2@illinois.edu},
  \texttt{dinasa2@illinois.edu},
  \texttt{jimeng@illinois.edu}
}

\begin{document}
\maketitle

\section{Abstract}

Medical dialogue systems (MDS) enhance patient-physician communication, improve healthcare accessibility, and reduce costs. However, acquiring suitable data to train these systems poses significant challenges. Privacy concerns prevent the use of real conversations, necessitating synthetic alternatives. Synthetic dialogue generation from publicly available clinical notes offers a promising solution to this issue, providing realistic data while safeguarding privacy. Our approach, \texttt{SynDial}, uses a single LLM iteratively with zero-shot prompting and a feedback loop to generate and refine high-quality synthetic dialogues. The feedback consists of weighted evaluation scores for similarity and extractiveness. The iterative process ensures dialogues meet predefined thresholds, achieving superior extractiveness as a result of the feedback loop. Additionally, evaluation shows that the generated dialogues excel in  factuality metric compared to the baselines and has comparable diversity scores with GPT4.

\section{Introduction}

Effective communication between patients and physicians is crucial for accurate diagnosis and treatment planning in healthcare. Medical dialogue systems (MDS) facilitate this communication by enabling inquiries beyond self-reports and providing automated diagnoses and recommendations. MDS help extend medical accessibility, enhance patient experiences, and reduce healthcare costs. However, privacy concerns restrict the use of real patient conversations for MDS training, necessitating the synthesis of dialogues. Clinical notes in Electronic Health Records (EHRs) are written documents that detail a patient's medical history, symptoms, diagnoses, and treatments during healthcare visits. These notes can be utilized to generate synthetic dialogues between patients and physicians to train MDS effectively, ensuring they are well-equipped to handle diverse healthcare scenarios.

In this work, our main motivation is to build a model that can generate high quality synthetic dialogue datasets from clinical notes with the ultimate goal to maintain the following: first, to adhere to HIPAA regulations and mitigate privacy risks associated with real patient data; second, to address the limited availability of benchmark datasets and facilitate the development and evaluation of healthcare dialogue systems; third, to overcome the lack of realistic multiturn dialogue datasets, crucial for modeling real-world healthcare conversations. 

To address these goals, we propose \texttt{SynDial}, a novel approach utilizing publicly available MTS-Dialogue and MIMIC datasets to generate synthetic medical dialogue data, with potential applications in other clinical notes datasets. Our approach employs a single LLM through zero-shot prompting to generate the dialogues, iteratively refining the output until high-quality dialogues are produced. During this iterative process, the generated dialogues are evaluated using initial thresholds for similarity and extractiveness to ensure they meet desired quality standards. These initial metrics are used solely for refining the dialogues during generation. This iterative process aims for high quality, differentiating our method from earlier work by ensuring the generated dialogues meet predefined thresholds. Additionally, the intrinsic and extrinsic evaluation results for generated dialogues from \texttt{SynDial} have shown superior results in extractiveness and factuality metrics compared to other baselines. The introduction of a feedback loop is motivated by the need to enhance the quality of the generated dialogues, addressing issues that arise without such iterative refinement.

\begin{figure*}[!htb]
  \centering  \includegraphics[width=\linewidth]{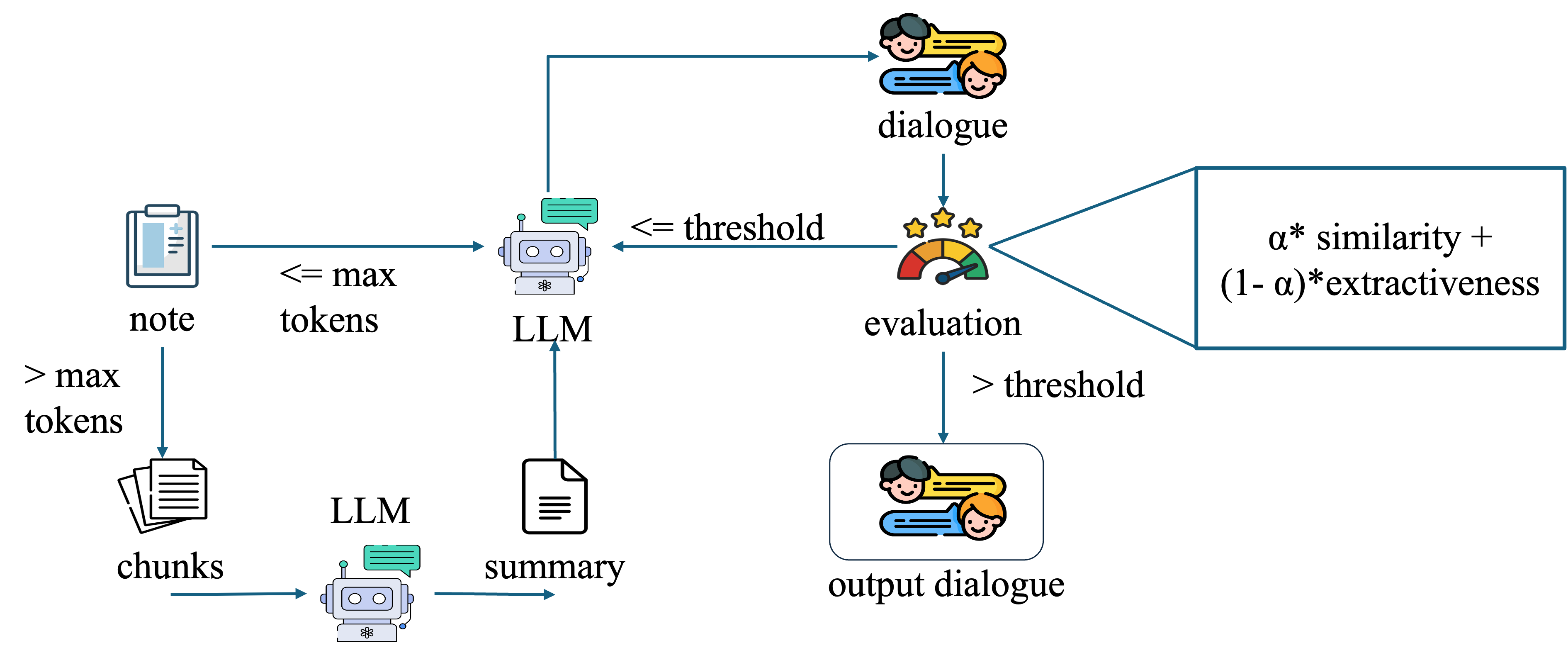}
  \caption{\texttt{SynDial} pipeline}
  \label{Fig:1}
\end{figure*}

\section{Related work}
\subsection{Medical Dialogue Systems}
Medical dialogue systems are designed to improve interactions between patients and healthcare providers by utilizing advanced natural language processing and machine learning techniques. These systems assist in diagnosing conditions, offering medical advice, and efficiently managing patient data. Various approaches to developing medical dialogue systems include deep reinforcement learning \cite{tang2016inquire, wei2018task, xu2019end}, pretrained language models \cite{varshney2023knowledge}, sequence-to-sequence models \cite{lin2019enhancing}, and meta-learning techniques \cite{lin2021graph}. Additionally, some methods incorporate knowledge graphs to enhance the system's understanding and performance \cite{lin2019enhancing, varshney2023knowledge}.

\subsection{Generating datasets for MDS training}
Recent advancements in MDS have significantly impacted the medical field. Some recent works have been done in generating synthetic patient-physician dialogue as an aim for providing a high quality of synthetic data that can be used for training MDS while protecting patient's privacy. The work in \cite{wang2023notechat}, proposed "NoteChat", a novel multi-agent framework using Large Language Models (LLMs) to generate synthetic patient-physician dialogues conditioned on clinical notes. NoteChat's structure includes Planning, Roleplay, and Polish modules. The Planning module focuses on knowledge organization to maintain medical logic. The Roleplay module employs two ChatGPT(GPT3.5Turbo) agents acting as patient and physician to produce interactive dialogues. Finally, the Polish module refines these dialogues to align closely with medical professional standards. The system was evaluated using the MTS-dialogue dataset on extensive intrinsic and extrinsic evaluation methods, and demonstrated improved performance over other models. Another study \cite{zeng2020meddialog}, introduces "MedDialog," a substantial collection of medical dialogues in both Chinese and English, forming the largest known dataset of its kind to date. MedDialog's approach to dialogue generation involves sequence-to-sequence modeling, utilizing advanced methods like BERT-GPT, to train models capable of producing contextually relevant and medically accurate conversations. The paper explores the effectiveness of these models and evaluates their performance on the MedDialog dataset, comparing them with other datasets to highlight their comprehensive coverage and diversity.

\section{Method} \label{sec: method}
In this section, we introduce our methodology, \texttt{SynDial}, which forms the core of our research. We will provide a detailed overview of the datasets employed and the evaluation metrics utilized to assess the efficacy of our approach in generating medical dialogues.

\subsection{\texttt{SynDial}}

The \texttt{SynDial} approach, as illustrated in Figure \ref{Fig:1}, begins by checking if the clinical note's length nearly exceeds the maximum token input for GPT-3.5 (if >4000). If it does, the note is first summarized by prompting GPT3.5 to makes it < 300 tokens. Subsequently, the clinical note is input into the LLM (GPT-3.5) to generate a dialogue between a patient and physician using zero-shot prompting. The generated dialogue is then initially evaluated and refined using the ROUGE-1 F1 metric for both similarity and extractiveness. Similarity measures how closely the generated dialogue matches a reference dialogue, while extractiveness assesses how much information in the dialogue is directly extracted from the clinical note.

The ROUGE-1 F1 score is calculated using the following equation:

\[
\text{ROUGE-1 F1} = 2 \cdot \frac{\text{precision} \cdot \text{recall}}{\text{precision} + \text{recall}}
\]

where precision is the fraction of relevant words retrieved from the generated dialogue, and recall is the fraction of relevant words retrieved from the reference dialogue.

The approach is flexible and can be tweaked using the parameter \(\alpha\), allowing for a balance between extractiveness and similarity. The combined score is calculated using the equation:

\[
\text{combined\_score} = (1-\alpha) \cdot \text{score\_extr}+ \alpha \cdot \text{score\_sim}
\]

If the dialogue meets predefined thresholds for the combined score, the process terminates, and the dialogue is saved. If not, a feedback loop sends the clinical note back to the LLM, along with the previously calculated similarity and extractiveness scores, to improve the dialogue. This loop continues until the dialogue meets the thresholds or until it has run three times. If it runs three times without meeting the thresholds, the best dialogue generated in terms of the combined score is selected. After evaluating the approach, the same process is applied for other datasets, focusing on extractiveness when ground-truth dialogues are not available.

\subsection{Datasets}

We utilize two datasets: MIMIC-IV \cite{johnson2023mimic} and MTS-Dialogue \cite{abacha2023empirical}. The MIMIC datasets are large publicly available electronic health records comprising 123,488 patients with 232,263 visits, averaging 1.88 visits per patient. We used 80 samples from the MIMIC-IV dataset for our experiments. Each patient typically has one to two visits, each corresponding to a discharge summary. These discharge summaries are used to generate synthetic dialogue datasets.

The MTS-Dialogue dataset contains clinical notes and corresponding ground truth conversations for each sample. It includes 1,701 pairs of dialogues and associated sections from clinical notes, with an overall of 15,969 turns, 18,406 sentences, and 241,685 words in dialogues, and summaries consisting of 5,870 sentences. For our experiments, we used the MTS-Dialogue test dataset containing 20 samples and the training dataset containing 1,200 samples, which were used for fine-tuning Llama2 models during extrinsic evaluation.

\subsection{Baselines}

\begin{itemize}
    \item GPT3.5 and GPT4
    both are using one detailed zero-shot instruction prompting to generate the dialogue between patient and physician \cite{wang2023notechat}. 
    \item NoteChat
    uses two LLMs (GPT3.5Turbo) where one acts as patient and the other as physician to generate a multi-turn conversation\cite{wang2023notechat}.
\end{itemize}

\subsection{Evaluation metrics} 
We evaluate our model and the baseline models based on both intrinsic and extrinsic evaluation metrics. 

\subsubsection{Intrinsic evaluation} \label{sec: intrinsic}

\begin{itemize}
    \item \textit{Similarity}: ROUGE-F1 is calculated to measure the similarity of the generated conversation and the ground-truth dialogues.
    \item \textit{Factuality}: Concept-Recall is calculated using GPT3.5 to extract medical concepts from model-generated dialogues and notes to get two corresponding concept lists and then calculate the overlap of medical concept.
    \item \textit{Extractiveness}: ROUGE-F1 of src->hypo assesses how much information in the dialogue is extracted directly from the clinical note.
    \item \textit{Diversity}: Self-BLEU calculates the variety in the generated patient and physician utterances.
\end{itemize}

\subsubsection{Extrinsic evaluation}
For extrinsic evaluation, we first finetuned a Llama-2 models on MTS-Dialogue training dataset and calculated extractiveness (Rouge 1 F1 score). We also finetuned another similar model with MTS-Dialogue training dataset augmented with MIMIC dataset generated by our method. We hope that adding our dataset will improve the score if tested on MTS-Dialogue test data on the Conversation2Note task.

\begin{table*}[!htb]
\centering
\begin{tabular}{@{}lrrrr@{}}
\toprule
                                  & \multicolumn{1}{l}{\textbf{Similarity}} & \multicolumn{1}{l}{\textbf{Extractiveness}} & \multicolumn{1}{l}{\textbf{Diversity}} & \textbf{Factuality} \\ \midrule
\textbf{GPT3.5}                  & 0.48                                  & 0.46                                      & 0.75                                 & 0.49              \\
\textbf{GPT4}                     & 0.53                                  & 0.50                                      & 0.74                                & 0.50              \\
\textbf{NoteChat}                 & \textbf{0.56}                         & 0.38                                      & \textbf{0.76}                        & 0.34              \\
\textbf{\texttt{SynDial}}            & 0.43                                  & \textbf{0.52}                             & 0.73                                 & \textbf{0.53}     \\
\bottomrule
\end{tabular}
\caption{Comparison among baselines on MTS-Dialogue data}
\label{tab: baselines}
\end{table*}

\section{Experiments and Results}
In this section, we describe the different experiments we have done including comparison with baselines for intrinsic evaluation, extrinsic evaluation, reward comparison, robustness check and cost comparisons.

\subsection{Comparison with baselines}
We compared our model on the MTS-Dialogue dataset with the baseline models. As done in the NoteChat \cite{wang2023notechat} paper, we compared four different intrinsic evaluation metrics across all the methods. The metrics (similarity, extractiveness, factuality, diversity) are described in Section \ref{sec: method}. Table \ref{tab: baselines} shows the comparison among all the methods across all the four intrinsic evaluation metrics. We can clearly see that \texttt{SynDial} is performing best for extractiveness and factuality. For our case (with MIMIC dataset), where we do not have the ground truth dialogues available for each clinical note, we find extractiveness and factuality as the most important evaluation metrics. Diversity if \texttt{SynDial} is comparable with GPT4 which is reasonable. 

\noindent For similarity and extractiveness we calculated Rouge1-F1 scores as described in section \ref{sec: intrinsic}. We utilize rouge-score 0.1.2 python package \cite{lin2004rouge} for calculating rouge scores. For factuality, we used gpt3.5 to first extract 2 sets of medical concepts from the note and the generated dialogue. Then we use porter stemmer stemming algorithm to process the medical concepts from both sets. Finally, we find the intersection of the sets and calculate recall. For diversity, we calculated self-bleu scores inspired by the github repository of \cite{zhu2018texygen}. We calculated extractiveness and factuality for a subset of 80 samples from the MIMIC IV dataset (Table \ref{tab:mimic}). For MIMIC IV data, we do not have the ground truth dialogues, so we can not calculate similarity with ground truth.

\begin{table}[]
\centering
\resizebox{\columnwidth}{!}{%
\begin{tabular}{@{}lrrr@{}}
\toprule
\textbf{}                    & \multicolumn{1}{l}{\textbf{Extractiveness}} & \textbf{Factuality} & \textbf{Diversity} \\ \midrule
\textbf{GPT3.5}              & 0.44                                        & 0.51                & \textbf{0.82}               \\
\textbf{\texttt{SynDial} }   & \textbf{0.49}                               & \textbf{0.52}       & 0.81               \\ \bottomrule
\end{tabular}%
}
\caption{Results on a subset of MIMIC data}
\label{tab:mimic}
\end{table}

\subsubsection{Extrinsic evaluation}
For extrinsic evaluation, we first finetuned a Llama-2-7b-hf model on the MTS-Dialogue training dataset (1200 samples) and calculated extractiveness (Rouge 1 F1 score). We also finetuned another Llama-2-7b-hf model with MTS-Dialogue training dataset augmented with MIMIC dataset generated by our method. We hope that adding our dataset will improve the score if tested on MTS-Dialogue test data. Table \ref{tab: extrinsic} shows the scores for extrinsic evaluation. Augmenting the MTS-Dialogue training dataset with synthetic MIMIC dialogues helped to elevate the Rouge 1 F1 score for Conversation2Note downstream task.

\subsection{Experiment for robustness check}
We run the same pipeline (Fig. \ref{Fig:1}) for three times and average the scores. We find that the scores are not too diverse with only a standard deviation of 0.007 for similarity rouge scores and a standard deviation of 0.003 for extractiveness rouge scores. This experiment shows that the scores are not random and are consistent enough.

\begin{table}[]
\begin{tabular}{@{}rrr@{}}
\toprule
                       & \textbf{Similarity} & \textbf{Extractiveness} \\ \midrule
\textbf{Experiment\#1} & 0.4269              & 0.5219                  \\
\textbf{Experiment\#2} & 0.4153              & 0.5208                  \\
\textbf{Experiment\#3} & 0.4282              & 0.5161                  \\
\textbf{avg}           & 0.4234              & 0.5196                  \\
\textbf{std}           & 0.0071              & 0.0030                  \\ \bottomrule
\end{tabular}
\caption{Robustness on MTS-Dialogue dataset}
\label{tab: robustness}
\end{table}
\subsection{Cost comparison}
\begin{figure}[!htb]
  \centering  \includegraphics[width=\linewidth]{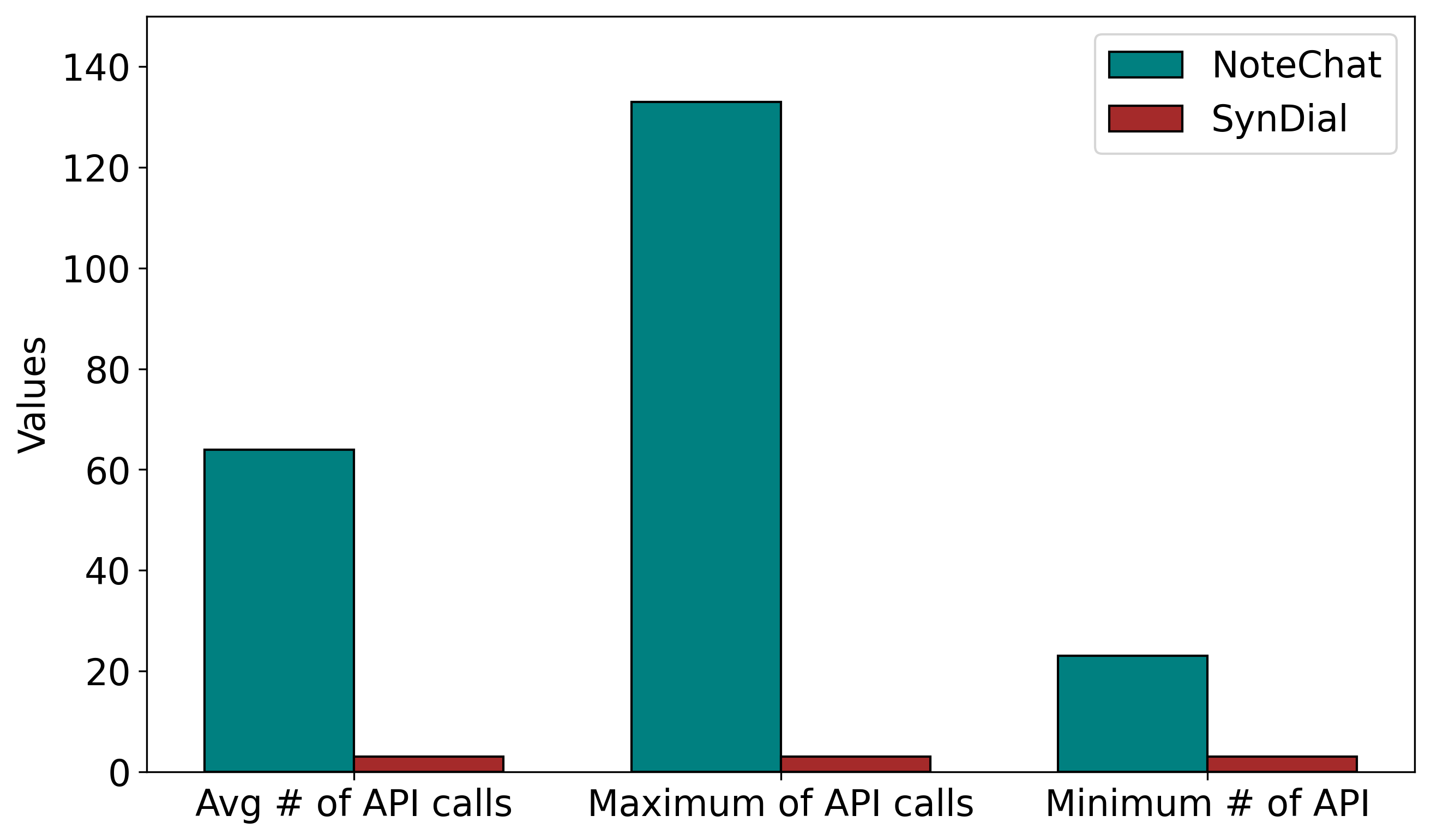}
  \caption{Cost comparison between NoteChat and \texttt{SynDial}}
  \label{Fig: cost}
\end{figure}

We also compared \texttt{SynDial} with NoteChat \cite{wang2023notechat} which is the state-of-the-art model that also generates dialogues from clinical notes. \texttt{SynDial} is far cheaper in terms of API calls compared to NoteChat (Figure \ref{Fig: cost}). Also, although NoteChat is better in terms of similarity to ground truth dialogues and diversity, we think it is definitely more important to have better extractiveness and factuality scores. 

\begin{table}[]
\centering
\begin{tabular}{@{}ll@{}}
\toprule
Training data  & Extractiveness  \\ \midrule
MTS           & 0.32526                                  \\
MTS + MIMIC   & \textbf{0.33297}                         \\ \bottomrule
\end{tabular}
\caption{Extrinsic evaluation on MTS-Dialogue test data for Conversation2Note task. }
\label{tab: extrinsic}
\end{table}

\section{Limitations and Future Work}
In the initial phase of our research, we utilized 80 samples from the MIMIC IV dataset to demonstrate performance. However, for the forthcoming paper, we plan to incorporate a larger sample size and integrate data from the MIMIC III dataset. Our hypothesis that patient history significantly aids in generating dialogues for current visits was refuted by experimental results. Hence, we aim to streamline the process by extracting pertinent information solely from the previous visit to inform prompts for subsequent visits. In future endeavors, we aspire to refine the feedback loop to provide more nuanced and reflective feedback, thereby enhancing the extractiveness of the generated dialogue.

\section{Conclusion}

In this paper, we presented \texttt{SynDial}, a novel approach for generating synthetic patient-physician dialogues from clinical notes using a single large language model (LLM) with a feedback loop mechanism. Our method addresses the challenges of data scarcity and privacy in training medical dialogue systems by leveraging publicly available clinical notes datasets such as MIMIC-IV and MTS-Dialogue. Through extensive experiments, we demonstrated that \texttt{SynDial} significantly outperforms existing baseline models in terms of extractiveness and factuality, making it a valuable tool for creating high-quality synthetic dialogue datasets.

Furthermore, we showed that our approach is cost-effective compared to the state-of-the-art models like NoteChat, which rely on multiple LLM instances. \texttt{SynDial} also provides consistent results across multiple runs, ensuring robustness in the generated dialogues. 

Future work will focus on scaling up the dataset size and incorporating more advanced feedback mechanisms to further enhance the quality of the generated dialogues. Additionally, we plan to explore the impact of integrating synthetic dialogues from multiple sources to improve the performance of downstream tasks, such as the Conversation2Note task.

Overall, \texttt{SynDial} offers a promising solution for advancing medical dialogue systems while maintaining patient privacy and reducing the dependency on real-world data.



\section{Conflicts of interest}
The authors have no competing interests to declare.

\section{Data availability}
The datasets used in this study are publicly available. The MIMIC-IV dataset can be accessed at \url{https://physionet.org/content/mimiciv/2.0/} \cite{johnson2023mimic}. The MTS-Dialogue dataset is available at \url{https://github.com/abachaa/MTS-Dialogue} \cite{abacha2023empirical}.

\bibliography{custom}
\bibliographystyle{acl_natbib}

\newpage
\appendix
\section{Appendix} \label{sec: appendix}
\subsection{Zero-Shot Dialogue Prompting}
The zero-shot prompting used in our approach for dialogue generation is as follows:

\texttt{prompt = f""" \\
Given the clinical note, write a conversation between the patient and the doctor from the clinical notes so that the main keywords are covered. \\
("the combined rouge score for both extractiveness and similarity for the previous dialog was " + \\
str(combined\_score) + " for this note " + row['note'] + \\
"generate a new one and try to improve the accuracy where the extractiveness should weigh"+(1-alpha)+"and the similarity should weigh"+alpha) \\
Dialogue: \\"""}

\subsection{Experiments with historical visits}
\begin{figure*}[!htb]
  \centering  \includegraphics[width=0.8\linewidth]{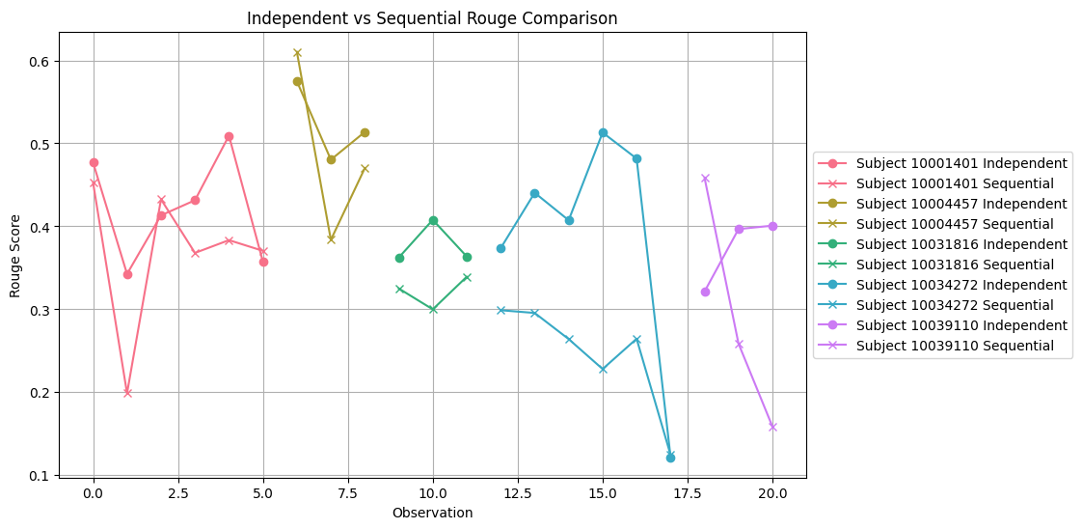}
  \caption{Including  vs not including previous visit's dialog and score in prompt. }
  \label{Fig:2}
\end{figure*}

We tried to incorporate information from previous visit of a patient in the prompt to help generating dialogue for that patient's current visit. However, we notice that the scores do not improve much. Figure
\ref{Fig:2} shows the score for 5 patients. Each color denotes a different patient. For each patient we plot the Rouge 1 F1 score for  generated dialogues independent of the previous visit (denoted with filled circles). We also plot the scores where for each visit (except visit 1 of each patient), the previous visit's dialogue is also used in the prompt and \texttt{SynDial}'s pipeline is used to generate dialogues. We see a decrease in overall scores for the second case (x marks in the graph). 

\subsection{Improvement in Extractiveness Scores}
\begin{figure*}[!htb]
  \centering  \includegraphics[width=0.8\linewidth]{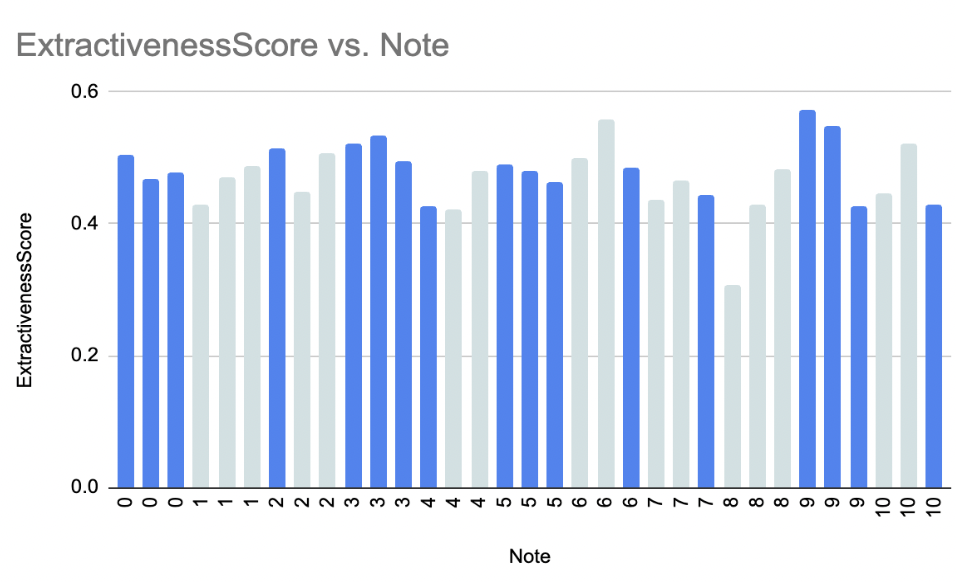}
  \caption{Improvement in Extractiveness Scores on MTS-Dialogue Dataset }
  \label{Fig:3}
\end{figure*}

\begin{figure*}[!htb]
  \centering  \includegraphics[width=0.8\linewidth]{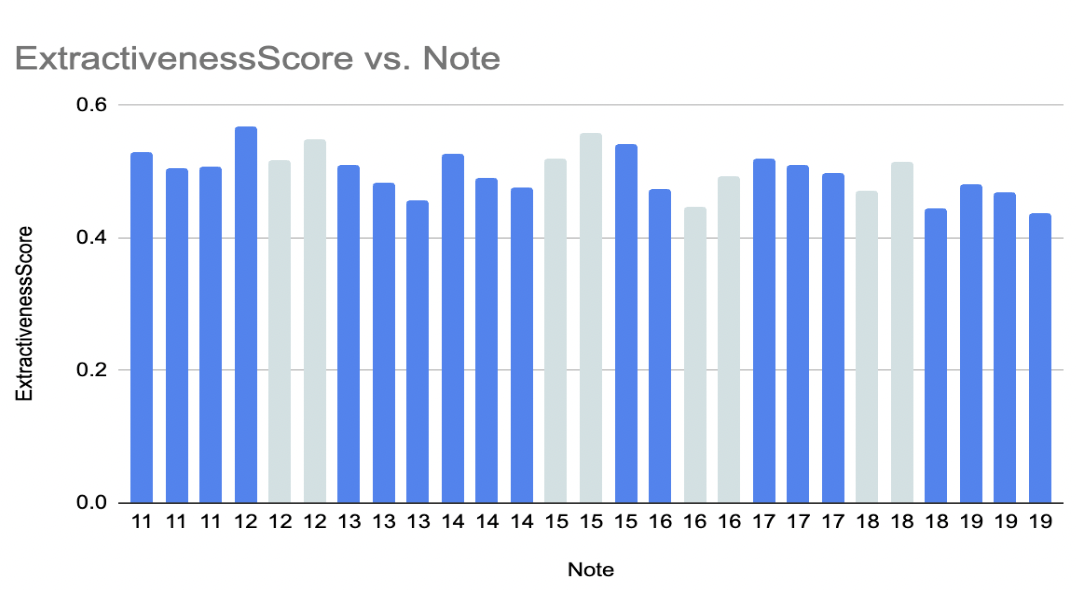}
  \caption{Improvement in Extractiveness Scores on MTS-Dialogue Dataset }
  \label{Fig:4}
\end{figure*}

We conducted a visual analysis to observe the changes in extractiveness scores within the MTS-Dialogue dataset across three iterations, as illustrated in Figures \ref{Fig:3} and \ref{Fig:4}. It was observed that, out of 20 clinical notes, 11 showed improvements in extractiveness scores across at least two consecutive iterations.

\subsection{Hyperparameter tuning}
Table \ref{tab:hyperparameter} shows the similarity and extractiveness scores when we tune the hyperparameter $\alpha$ which balances the priority of similarity and extractiveness. When $\alpha$ is 0, then full weight goes to extractiveness. If $\alpha$ is 1, the similarity score is calculated only as reward. When there are no ground truth dialogues, we need to select $alpha = 0$. However, if there are both the notes and corresponding ground truth dialogues the value for $\alpha$ can be chosen based of the need of the users. From Table \ref{tab:hyperparameter} we find $\alpha = 0.1$ is good for both similarity and extractiveness. 

\begin{table}[]
\centering
\begin{tabular}{@{}rrr@{}}
\toprule
\multicolumn{1}{l}{$\alpha$} & \multicolumn{1}{l}{\textbf{Similarity}} & \multicolumn{1}{l}{\textbf{Extractiveness}} \\ \midrule
0                         & 0.4153                         & 0.5208                             \\
0.1                       & 0.4288                         & 0.502                              \\
0.4                       & 0.4126                         & 0.4821                             \\
0.5                       & 0.42                           & 0.4893                             \\
0.7                       & 0.4347                         & 0.4909                             \\
1                         & 0.5055                         & 0.4831                             \\ \bottomrule
\end{tabular}
\caption{Hyperparameter tuning for $\alpha$}
\label{tab:hyperparameter}
\end{table}

\end{document}